\documentclass[10pt,twocolumn,letterpaper]{article}

\usepackage{cvpr}              

\usepackage{graphicx}
\usepackage{amsmath}
\usepackage{amssymb}
\usepackage{booktabs}
\usepackage{tabularray}
\usepackage{siunitx,booktabs}
\usepackage{tabularx, makecell, booktabs}

\usepackage[pagebackref,breaklinks,colorlinks]{hyperref}

\usepackage[capitalize]{cleveref}
\crefname{section}{Sec.}{Secs.}
\Crefname{section}{Section}{Sections}
\Crefname{table}{Table}{Tables}
\crefname{table}{Tab.}{Tabs.}

\def\cvprPaperID{27} 
\def\confName{CVPRW}
\def\confYear{2025}

\begin{document}

\title{EvtSlowTV - A Large and Diverse Dataset for Event-Based Depth Estimation}

\author{Sadiq Macaulay\\
University of Surrey \\
{\tt\small s.macaulay@surrey.ac.uk}
\and
Nimet Kaygusuz\\
University of Surrey\\
{\tt\small nimet.kaygusuz@surrey.ac.uk}
\and
Simon Hadfield\\
University of Surrey\\
{\tt\small s.hadfield@surrey.ac.uk}
}

\maketitle

\begin{abstract}
Event cameras, with their high dynamic range (HDR) and low latency, offer a promising alternative for robust depth estimation in challenging environments. However, many event-based depth estimation approaches are constrained by small-scale annotated datasets, limiting their generalizability to real-world scenarios. To bridge this gap, we introduce EvtSlowTV, a large-scale event camera dataset curated from publicly available YouTube footage, which contains more than 13B events across various environmental conditions and motions, including seasonal hiking, flying, scenic driving, and underwater exploration. EvtSlowTV is an order of magnitude larger than existing event datasets, providing an unconstrained, naturalistic setting for event-based depth learning. This work shows the suitability of EvtSlowTV for a self-supervised learning framework to capitalise on the HDR potential of raw event streams. We further demonstrate that training with EvtSlowTV enhances the model's ability to generalise to complex scenes and motions. Our approach removes the need for frame-based annotations and preserves the asynchronous nature of event data. 
The EvtSlowTV dataset and tools can be downloaded at $<$URL removed for anonymous review$>$.



\end{abstract}

\section{Introduction}
\label{sec:intro}


    \begin{figure*}[tb]
        \centering
        \includegraphics[height=8cm, width=1.05\textwidth, trim=3.1cm  2.0cm 1.0cm 2.0cm,  clip]{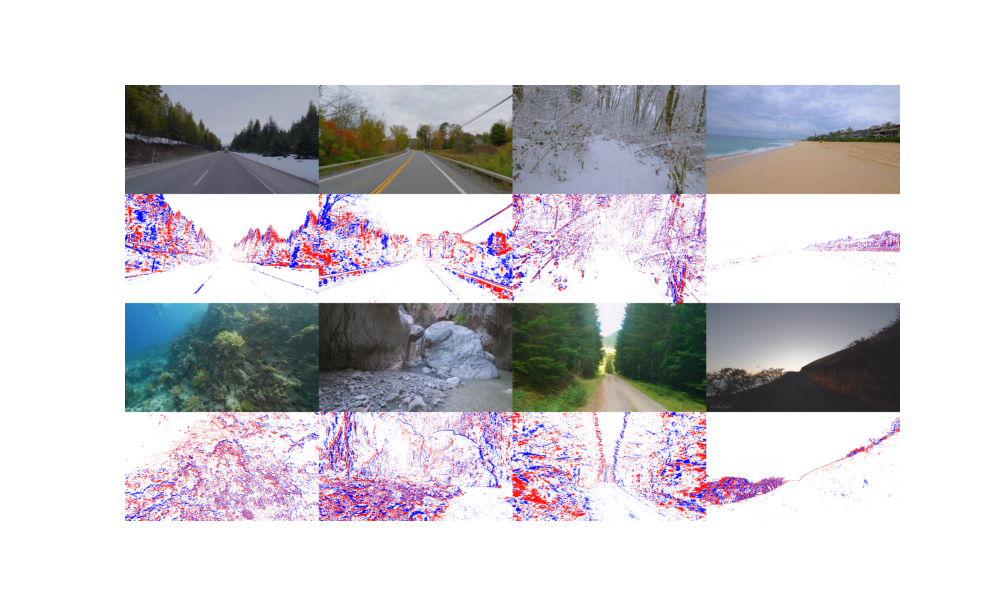}
        \caption{ Sample images and events extracted from the proposed dataset, featuring diverse scenes of hiking, driving, drones and scuba diving. The event dataset was generated from 45 videos curated from YouTube, totalling $\approx$2M frames.}
        \label{fig:samples}
    \end{figure*}

Event cameras, also known as neuromorphic vision sensors, offer a paradigm shift in computer vision by providing asynchronous, high-temporal-resolution, and high-dynamic-range (HDR) imaging capabilities. Unlike conventional frame-based cameras, which capture images at fixed frame rates, event cameras respond to changes in brightness at the pixel level, enabling low-latency perception even in challenging lighting conditions \cite{gallego2022event}. These unique characteristics have increased interest in leveraging event cameras for depth estimation and 3D scene understanding in robotics, autonomous vehicles, augmented reality and medical imaging applications \cite{Xin2024}. Frame-based stereo vision, monocular depth prediction, and structure-from-motion (SfM) are widely explored techniques for estimating depth maps \cite{zhu2021eventgan}. Sadly, these methods rely on well-lit high-contrast images, making them unsuitable for low-light, high-speed or high-dynamic range scenarios. In contrast, event cameras provide a promising alternative for depth estimation, particularly for low-latency applications \cite{zheng2023deep}. Despite these advantages, their asynchronous spatiotemporal data representations often require synchronized annotation with RGB, LiDAR, or stereo sensor data \cite{gehrig2021dsec, qu2023selfodom}, to apply deep learning methods. This diminishes the asynchronous nature of event cameras and limits their real-time performance \cite{chen2024data}. While self-supervised learning-based methods may be an alternative approach, the potential of this solution has not yet been optimally explored. This is affected by the challenges faced with large-scale, high-quality data collection for event-based vision \cite{gehrig2021dsec}. 



To address the problem, we introduce the EvtSlowTV dataset, a large-scale event camera dataset curated from real-world video footage to provide unconstrained, naturalistic depth training data (see figure \ref{fig:samples}). The dataset contains over 10B events, covering diverse real-world environments and motion patterns including seasonal hiking trails, drone footage, urban driving, and underwater exploration \ref{tab:event_camera_ds}. Unlike previous datasets that rely on controlled or simulated scenes, EvtSlowTV captures real-world depth variations in naturalistic settings, improving generalization for deep learning models.

This work presents the following contributions:
\begin{enumerate}
    \item A Large-Scale Dataset for Event-Based Depth Estimation: EvtSlowTV is an order of magnitude larger than existing event-based datasets, providing unconstrained depth variation and high-diversity scenarios for training robust models.
    \item Self-Supervised Depth Learning Framework Without External Sensors: We propose a self-supervised learning framework that removes the dependency on RGB, LiDAR, or stereo supervision and preserves the asynchronous nature of event cameras.
    \item Validation of the Event Dataset Generation for Improved Generalization and Accuracy: We showed that by leveraging on the large-scale and varied sequences in the EvtSlowTV, depth map can be learned directly from the spatiotemporal event representation without needing auxiliary sensor annotation.
\end{enumerate}


\section{Related Work}
Event-based cameras have gained significant attention recently due to their low latency, high dynamic range (HDR), and low power consumption, making them ideal for robotics and computer vision applications \cite{gallego2022event}. These advantages allow event cameras to excel where traditional frame-based cameras struggle, such as in low-light environments and high-speed motion scenes. However, challenges persist in data processing, synchronization, and the availability of large-scale datasets for training deep learning models \cite{li2024blinkflowdatasetpushlimits, ghosh2024eventstereo}. \\

\noindent
\textit{A. Event-Based Depth Estimation}\\
 Depth estimation using event cameras has been explored through stereo vision \cite{zhang2024ematchunifiedframeworkeventbased, Chen_2023}, monocular depth \cite{hidalgo2020monocular, Ilani2022}, and optical flow-based \cite{yang2024spiking} methods that can be extended to depth estimation tasks. Stereo event-based depth estimation is a widely studied approach, where two synchronized event cameras are used to estimate disparity maps, similar to traditional stereo vision methods \cite{ghosh2024eventstereo}. The DSEC dataset introduced by Gehrig et al. \cite{gehrig2021dsec} has been a critical dataset for stereo event-based disparity estimation, enabling deep learning models to train on event-based disparity maps. Monocular event-based depth estimation remains a greater challenge due to the lack of explicit disparity information. Many monocular and flow-based approaches introduce a secondary supporting sensor (either RGB or depth) \cite{Soliman_2024, niu2024imuaid, Chen_2023}. These auxiliary sensors operate in a synchronous frame-based manner, meaning the resulting techniques often estimate depth at fixed discrete intervals.\\

\noindent
\textit{B. Self-Supervised Event-Based Depth Estimation}\\
Supervised learning approaches for event-based depth estimation generally depend on annotated datasets that contain accurate ground-truth depth maps. However, acquiring such datasets is challenging and costly, as it typically involves the integration of external sensors, like LiDAR or structured-light devices \cite{zhu2018mvsec, gehrig2021dsec}. Additionally, reliance on external frame-based ground-truth depth undermines the inherent advantages of event cameras, specifically their asynchronous, high temporal-resolution characteristics, as ground truth is usually available only at fixed, discrete intervals \cite{gehrig2021eventsframe, hidalgo2020monocular}.
To mitigate these issues, self-supervised approaches have recently emerged as promising alternatives. These methods infer depth by leveraging motion constraints, unsupervised photometric losses, and contrastive learning paradigms, eliminating the need for explicitly labeled depth data \cite{meng2024learning, karmokar2024secrets}. Furthermore, self-supervised techniques preserve the event camera’s inherent asynchronous capabilities, avoiding frame-based supervision or dependency on auxiliary sensors.

Recent progress in event-based self-supervised learning demonstrates performance comparable to traditional supervised approaches, significantly reducing reliance on manually annotated depth maps. For example, Hagenaars et al. \cite{hagenaars2024self} presented a self-supervised monocular depth estimation framework capable of running in real-time on-device, highlighting the practical feasibility and efficiency of self-supervised event-based depth estimation for real-world applications.

Unfortunately, despite these promising advances, the limited diversity and scale of existing event-based datasets remain significant barriers. These limitations motivate the introduction of datasets such as EvtSlowTV, which provides an order of magnitude more data than existing benchmarks and expands the data domain from simple driving scenarios to unconstrained nature, indoor, underwater and drone footage. This offers improved opportunities for training robust, self-supervised event-based depth estimation models.

\noindent
\textit{C. Event Camera Datasets for Depth Estimation}

Large-scale event camera datasets play a critical role in training robust deep-learning models. Traditional datasets such as MVSEC \cite{zhu2018mvsec} and DSEC 
\cite{gehrig2021dsec} provide event streams paired with depth maps, but their limited scene diversity and small dataset size restrict their use for generalizable models. More recent synthetic datasets, such as EventScape \cite{gehrig2021eventsframe}, attempt to bridge this gap by generating event-based data from simulated environments. However, synthetic datasets often fail to generalize to real-world data due to domain adaptation issues. We introduce EvtSlowTV, a large-scale dataset that captures naturalistic depth variations from real-world YouTube sequences to address this gap. Unlike prior datasets, EvtSlowTV provides an order of magnitude more event data from diverse scenarios (see table \ref{tab:event_camera_ds}), enabling self-supervised learning at scale while preserving the asynchronous nature of event cameras.


\section{EvtSlowTV Dataset Generation}
\label{sec:event_ds}

\begin{table*}[t!]
    \centering
    \caption{Comparison of Event Camera Dataset Benchmarks}
    {\footnotesize
    \begin{tabular}{l*{9}{c}} 
      \toprule
      Dataset & 
      \multicolumn{9}{c}{Categories} \\ 
      \cmidrule(lr){2-10}
      & Natural & Indoor & Outdoor & Hiking & Driving & Flying & Depth & No. Sequence & Total Durations (min) \\             
      \midrule
       MVSEC \cite{zhu2018mvsec}                  & Yes & Yes & Yes & No  & Yes & Yes & Yes & 5 & 160 \\[2mm]
       EventScape \cite{gehrig2021eventsframe}    & No  & Yes & Yes & No  & Yes & No  & Yes & 758 & 800 \\[2mm]
       DESEC \cite{gehrig2021dsec}                & Yes & No  & Yes & No  & Yes & No  & Yes & 53 & 500 \\[2mm]
       DDD17 \cite{binas2017ddd17}                & Yes & No  & Yes & No  & Yes & No  & No  & 39 & 35 \\[2mm]
       DDD20 \cite{hu2020ddd20}                   & Yes & No  & Yes & No  & Yes & No  & No  & 12 & 600 \\[2mm]
       EvtSlowTV (Ours)                                       & No  & Yes & Yes & Yes & Yes & Yes  & No  & 40 & 9000 \\[2mm]
      \bottomrule
    \end{tabular}
    }
    \label{tab:event_camera_ds}
\end{table*}

We introduce EvtSlowTV, a large-scale synthetic event dataset generated from real-world video sequences. It leverages adaptive event sampling to create high-fidelity event representations while reducing the acquisition constraints for capturing scalable data from real cameras. Our dataset generation is inspired by SlowTV, a publicly available video dataset that captures long-duration real-world sequences across diverse environments, including indoor, outdoor, and dynamic scenes for RGB-based monocular depth estimation \cite{spencer2023kickrelaxlearning}. We implemented an automated toolchain for converting videos into event streams using ESIM, an Open-Event Camera Simulator that models per-pixel intensity changes between consecutive frames \cite{rebecq2018esim}. This method employs an adaptive rendering strategy, where frames are sampled only when significant changes in visual signal occur, ensuring event sparsity and temporal precision. 

\subsection{Adaptive Frame Sampling}
 Unlike conventional uniform frame sampling, which can introduce redundancy in static regions and miss details in dynamic areas, adaptive sampling adjusts the sampling rate based on brightness changes and scene motion. The EvtSlowTV dataset leverages an adaptive sampling strategy to generate high-fidelity event streams from real-world video sequences while preserving critical motion and brightness variation cues. 

  To achieve adaptive sampling of frames from a sequence, the logarithmic irradiance (brightness) gradient is computed across spatial and temporal dimensions, ensuring consistency across different scenes. For each frame time $t_k$ and position $x, y$, the system estimates the brightness change rate $\frac{\partial L(x, y; t_k)}{\partial t}$ that describe pixel-wise intensity evolution. The consecutive frames time $t_{k+1}$ in a sequence is determined by:


\begin{equation}
    t_{k+1} = t_k + \left(C \left| \max_{(x, y) \in \Omega} \frac{\partial L(x, y; t_k)}{\partial t} \right|^{-1} \right)
\end{equation}
where $\Omega$ is the domain of 2D pixel positions. This means that if the brightness changes rapidly in certain regions, the frame sampling rate increases to ensure fine-grained event generation. Conversely, in static regions with minimal brightness variations or motion, the algorithm reduces the sampling rate to optimize computational efficiency. This behavior mimics the asynchronous properties of how events are being triggered by relative changes in brightness at a pixel from real event cameras. 

\subsection{Events From SlowTV}
The implementation begins with frame preprocessing, where frames from each sample in the SlowTV dataset are extracted, normalized, and analyzed for brightness changes separately. From the selected frames of the adaptive sampling, events are generated by detecting per-pixel intensity changes between the frames. 

For each pixel location in image space $(x, y)$ at time $t_k$, we compute the log-intensity change $L(x, y)_{t_k}$ based on the difference between the current frame and the last frame where an event was generated. An event is triggered when the absolute difference between the current log intensity and the last log intensity at which an event was generated exceeds a predefined contrast threshold $C$:

\begin{equation}
    \left| L(x, y)_{t_k} - L(x, y)_{t_{k-\tau}} \right| \geq C 
\end{equation}
where $\tau$ is the number of frames since the last event was triggered at pixel $x,y$. Each event is represented as an array $\mathbf{e} \doteq \left(x, y, t, p\right)$ combining a pixel location $x$ and $y$, timestamp $t$, and polarity $p \doteq [-1, 1]$ that indicates an increase or decrease in intensity in the respective pixel location. 

\begin{equation}
    p =
    \begin{cases}
    +1, & L(x, y)_{t_i} - L(x, y)_{t_{i-\tau}} \geq C \\
    -1, & L(x, y)_{t_i} - L(x, y)_{t_{i-\tau}} \leq -C
    \end{cases}
\end{equation}

From each video sample, we generated an array of events $E \doteq \left\{\mathbf{e_i}\right\}_{i=1}^{N}$ where $N$ is the total number of events triggered in the sequence. These event sets are provided in their original form as the EvtSlowTV dataset along with the tools to include additional videos or to process the events into other input formats (e.g. quantized event tensors).

\section{Event-Based Self-Supervised Depth Estimation}
Our event-driven monocular depth framework is driven by a simple observation. Events occur primarily at the boundaries of moving objects over a period of change. This means that if a scene is static and the camera moves, the edges remain in the same 3D locations but undergo apparent 2D motion based on camera movement \cite{Shiba24pami}. If we estimate the continuous camera motion and scene depth throughout the sequence, we can backproject the observed 2D events to a unified 3D point cloud. If our depth and pose estimates are accurate, all back-projected events should align precisely with object boundaries in 3D, This in turn means that the reprojection of the event 3D point cloud should result in a sharp and clean edge map in 2D \cite{zheng2023deep}. To train our network in a self-supervised manner, we optimize a contrast maximization loss function based on multi-view photometric consistency. This loss function enforces the correct alignment of back-projected events, ensuring that depth and camera motion are inferred in a geometrically consistent manner \cite{ZhouEventBasedStereo}. In the next section, we demonstrate a self-supervised and purely event-driven monocular depth estimation framework exploiting the EvtSlowTV dataset.

\subsection{Event-Based Depth-Pose Network}
\label{sec:event_net}
Raw event streams contain sparse, asynchronous data, making direct processing challenging. To enable learning-based depth estimation, we extract a subset of the event data $E_{t \to t^\prime} \subseteq E$ into a structured representation. $E_{t \to t^\prime}$ is aggregated from a reference time stamp $t$ over a duration up to $t^\prime$. For our implementation, we assumed a duration of 0.1665 seconds, corresponding to five consecutive frames from the standard 30 FPS camera frame rate. Event volumes are formed using linearly weighted
accumulations into 5 bins of the normalized timestamps. This creates a spatiotemporal tensor representation $E_{vol}$.

\begin{equation}
    E_{vol}(x, y, t) = \sum_{i}^{N} p_i\delta(x - x_i, y - y_i) \max\left(0, 1 - |t-t_i|\right)
\end{equation}
where the normalized timestamp is given as $t_i = \lfloor \frac{B-1}{t_N - t_1}\left(t_i-t_0\right) \rfloor$, $B$ is the number of bins and $\delta$ is the Dirac delta function. This event volume is used as input to the depth estimation network, it effectively encodes spatiotemporal motion cues, which are crucial for depth estimation with a self-supervised network model \cite{zheng2023deep}.

To estimate depth and camera pose, we employ a skip-connection encoder-decoder network $f_{\Theta}$, which extracts and refines hierarchical motion and structural features. The encoder compresses the multi-scale event representation into a latent feature space, capturing high-level motion cues. The decoder then reconstructs depth predictions while maintaining spatial consistency, aided by skip connections that preserve fine-grained details by allowing information to flow directly from the encoder’s early layers. This network architecture ensures that local structural details and global motion information are optimally leveraged, leading to more robust predictions \cite{LiuEventBasedDepth}.

\begin{equation}
    (\hat{\mathcal{D}}, \mathbf{T}) = f_{\Theta}(E_{vol} ) 
\end{equation}

The depth and camera pose estimation are formulated to enable geometric transformations of event-based features over time, where the estimated depth $\hat{\mathcal{D}}(x, y)$ can be used to retrieve the 3D world coordinate $\mathcal{P}$ using the camera intrinsic $K$ given as:

\begin{equation} 
    P_i = D(\bar{\mathbf{e}}_i) K^{-1} \bar{\mathbf{e}}_i
\end{equation}
where $\bar{\mathbf{e}}_i$ is the homogeneous 2D co-ordinates of event $i$. The predicted camera pose provides the transformation matrix $\mathbf{T} = \begin{bmatrix} \mathbf{R} & \mathbf{t} \end{bmatrix}$, which contains the rotational matrix $\mathbf{R}$ and translational vector $\mathbf{t}$ component of the camera motion between consecutive event frames $t \to t^\prime$.

The estimated pose allows us to transform points from one frame timestamp to another in a consistent manner given as;

\begin{equation} 
    \hat{\mathbf{E}}_{t \to t^\prime} = \mathbf{T} \otimes \mathcal{P}
\end{equation}
ensuring that all back-projected events align along real-world object boundaries. By predicting depth and pose jointly, the network reinforces the temporal coherence of the event from different timestamps to improve prediction accuracy.

\subsection{Contrast Maximization Loss}
To learn depth and pose prediction, we employ a Contrast Maximization (CM) loss, which enforces the temporal alignment of events to ensure accurate motion compensation. The fundamental idea behind CM is that events originate from the same scene edges but at different pixel locations and timestamps with respect to the camera motion. If depth and camera pose are accurately estimated, all events should be warped into a common reference frame, producing a sharp and high-contrast event image. This approach leverages the natural properties of event data, where maximizing the contrast in the accumulated events enhances the accuracy of motion estimation \cite{Shiba24pami}. 

We define an Image of Warped Events (IWE) by accumulating the wrapped events into an image frame:

\begin{equation} 
    I(x, y) = \sum_{i}^{N} p_i \delta(x - \hat{{x}}_i, y - \hat{y}_i) 
\end{equation}
where $\hat{x}_i$ and $\hat{y}_i$ pixel location of warped 
events in $\hat{\mathbf{E}}_{t \to t^\prime}$
To enforce the alignment of back-projected events, we maximize the variance of the IWE, leading to the contrast loss function:

\begin{equation} 
    \mathcal{L}_{contrast} = - \frac{1}{|\Omega|} \int_{\Omega} \left( I(x) - \mu_I \right)^2 dx 
\end{equation}
where $\Omega$ is the spatial domain and $\mu_I$ is the mean intensity of the warped event image. This loss function is minimized when perfectly aligned events result in high-contrast event reconstructions. By optimizing depth and pose jointly using contrast maximization, our model learns to produce sharp and geometrically consistent depth maps, making it well-suited for self-supervised event-based vision tasks without relying on frame-based auxiliary sensors or ground truth labels.


\subsection{Depth Optimization Losses}
    \begin{figure}[tb]
        \centering
        \resizebox{\columnwidth}{!}{%
        \includegraphics[scale=0.5,trim=0.0cm 0.0cm 5.0cm 0.0cm,  clip]{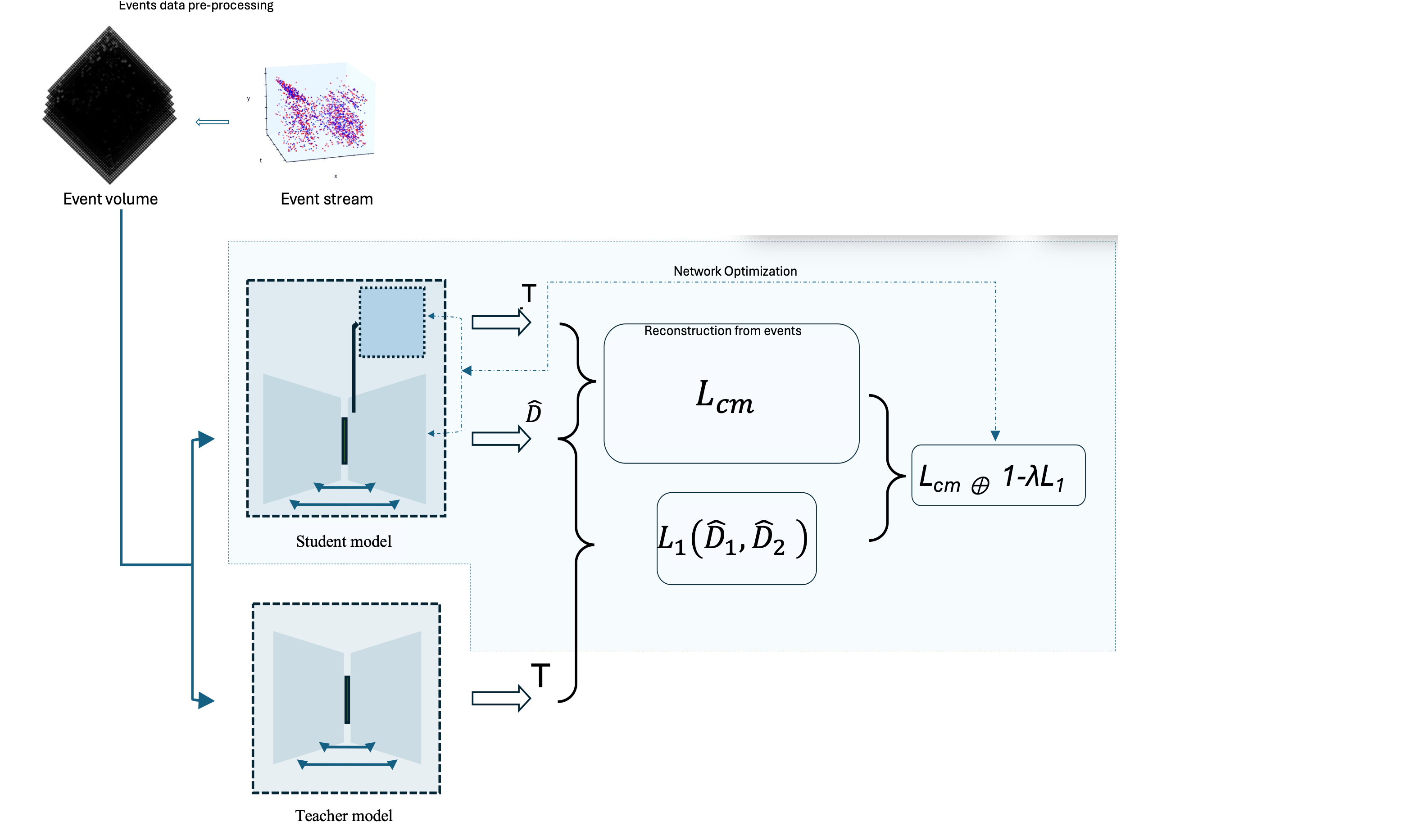}
        }
        \caption{The Teacher-Student training strategy adopted for self-supervised depth estimation model.}
        \label{fig:net}
    \end{figure}
The proposed self-supervised depth estimation model employs a student-teacher learning framework. This training strategy enables the transfer of knowledge from a more robust and stable supervised teacher model, while also fine-tuning on data without ground-truth supervision. This approach helps the student model adapt to the high temporal resolution and sparse nature of event data while ensuring better generalization across diverse motion scenarios. This allows us to define different loss functions to enforce specific learning paradigms to the model. To ensure the depth models are invariant to varying depth scales we enforce a scale-invariant loss function $\mathcal{L}_{si}$. This allows the model to learn relative depth differences consistently across sequences, and prevents the model from being biased toward specific depth ranges \cite{eigen2014depthmap}. 
\begin{equation}
    \mathcal{L}_ {si} = \frac{1}{n}\sum_\mathbf{u} \left(\mathcal{R}(\mathbf{u})\right)^2 - 
    \frac{1}{n^{2}}\left(\sum_\mathbf{u} \mathcal{R}(\mathbf{u})\right)^{2} ,
\end{equation}
where $n$ is the number of valid ground truth pixels $\mathbf{u}$ and $\mathcal{R}(\mathbf{u})$ is the depth error given as $\left| \hat{\mathcal{D}} - \mathcal{D}\right|$,

Since events are often observed along the boundaries of objects in the scene, it is important to encourage the model to learn structural details. By applying a scale-invariant gradient matching loss function $\mathcal{L}_{grad}$, the model can learn depth consistency and discontinuities \cite{li2018megadepth}.

    \begin{table*}[ht!]
      \centering
      \caption{Comparing depth estimation from different methods and reporting performance metrics on MVSEC \textit{indoor\_flying} sequences.}
      {\footnotesize
      \begin{tabular}{l*{6}{S[table-format=2.5]}}
          \toprule
          Methods & 
          \multicolumn{2}{c}{MVSEC flying1} & 
          \multicolumn{2}{c}{MVSEC flying2} & 
          \multicolumn{2}{c}{MVSEC flying3} \\ 

          \cmidrule(lr){2-3}
          \cmidrule(lr){4-5}
          \cmidrule(lr){6-7}
          
          & {Abs mean} $\downarrow$ & {rms\_log} $\downarrow$ 
          & {Abs mean} $\downarrow$ & {rms\_log} $\downarrow$
          & {Abs mean} $\downarrow$ & {rms\_log} $\downarrow$ \\
         
           \midrule
           EMVS \cite{Rebecq18ijcv}         &0.3937            &0.2210           &0.3142              &0.2180          &0.3054           &0.1827            \\[2mm] 
           ESVO \cite{ZhouEventBasedStereo} &0.2339            &0.1672           &0.2042              &0.1835          &0.2429           &0.1754            \\[2mm] 
           Ghosh et al.\cite{Ghosh_2022}    &0.2253            &\textbf{0.1411}  &0.1820              &\textbf{0.1359} &0.1949           &\textbf{0.1201}   \\[2mm] 
           Ours                             &\textbf{0.1887}   &0.5675           &\textbf{0.1659}     &0.4708          &\textbf{0.1882}           &0.4735            \\[2mm] 
           
          \bottomrule
      \end{tabular}}
      \label{tab:performance}
    \end{table*}

\begin{equation}
    \mathcal{L}_{grad}=\frac{1}{n} \sum_s \sum_{\mathbf{u}}\left|\nabla_x \mathcal{R}^s(\mathbf{u})\right|+\left|\nabla_y \mathcal{R}^s(\mathbf{u})\right| .
\end{equation}
where $\nabla_x$ and $\nabla_y$ are Sobel operators to extract edges along the $x$ and $y$ direction and $s$ indicates the scales. 

The two loss functions described above are dependent on ground truth depth annotations and can both be applied in a weighted combination \cite{gehrig2021eventsframe} to train the supervised teacher model.
\begin{equation}
    \mathcal{L}_{teacher} = \mathcal{L}_{si} + \lambda\mathcal{L}_{grad}
\end{equation}

The transfer learning strategy allows the student model prediction during training to be enhanced by mimicking the qualities of the teacher model prediction. This is made possible by combining the CM loss function with the $\mathcal{L}_1$ loss between the prediction from the frozen teacher model and the main student network model. This is given by:   
\begin{equation}
    \mathcal{L}_{student} = \mathcal{L}_{contrast} + (1-\lambda)\mathcal{L}_1(\hat{\mathcal{D}}_{1}, \hat{\mathcal{D}}_{2})
\end{equation}
where $\hat{\mathcal{D}}_{1}$ and $\hat{\mathcal{D}}_{2}$ are depth predictions from the teacher and student model, respectively while $\lambda = 0.6$ is the weight value. This enables the supervised teacher to act as a stabilizing influence during self-supervised training.

\section{Evaluation}

    \label{sec:results}
    We conducted extensive experiments to showcase the value of EvtSlowTV for self-supervised depth estimation, and the benefits of large-scale and varied data domains. Our proposed model, although simple, exhibits remarkable performance comparable to the SOTA methods.

\subsection{Experiment Set-Up}
    
To evaluate the effectiveness of the EvtSlowTV we implemented our depth network and train the final model in a self-supervised manner on this dataset. We employed a teacher-student training framework to train an event-based SSL model. To achieve this, we initially trained the teacher network in a supervised manner on the MVSEC dataset \cite{zhu2018mvsec}. The first 80\% of the duration of most sequences featured in this dataset (except the \textit{indoor\_flying}) was used for training, and the last 20\% for testing. The proposed models are implemented in PyTorch \cite{paszke2019pytorch}. Each model was optimally trained until convergence. However, we maintain a learning rate of $10^{-4}$ with a weight decay of $10^{-3}$ using AdamW\cite{loshchilov2019decoupled} and a batch size of 16 on a single NVIDIA GeForce RTX 3060 (12GB).

\subsection{Comparison to Supervised-Learning}
To assess the effectiveness of our proposed depth estimation model that has been trained on the EvtSlowTV, we compared our prediction errors with the state-of-the-art. For direct comparison, we tested our model on the \textit{indoor\_flying} sequence in the MVSEC dataset using the widely accepted absolute mean depth and log-mean-square (rms\_log) error metrics for depth estimation (see Table \ref{tab:performance}). The test samples make up about 200 second-long sequence and 4000 frames of ground-truth depth map with a maximum depth distance of 8.40 meters.

    \begin{table*}[!t]
        \centering%
        \caption{\textbf{Ablation}; Comparing mean square error of different training techniques over varied distances. It includes prediction error from the supervised training $EvtSSL$, finetuning $EvtSSL \to EvtSSL$ and student teacher model $EvtSSL \to \overline{EvtSSL}$ training setup.}
        \begin{tabular}{l|c|*{3}{c}}
            \specialrule{1.5pt}{1pt}{1pt}
            \toprule
            \thead{Sequence} 
            & \thead{Distance (meters)} 
            & \thead{$EvtSL$} 
            & \thead{$EvtSL \to EvtSSL$} 
            & \thead{$EvtSL \to \overline{EvtSSL}$} \\

            \midrule
            
            outdoor day1    &0 - 10     &0.4184       &0.2571       &0.2570 \\
                            &0 - 20     &0.4329       &0.3004       &0.3003 \\
                            &0 - 30     &0.3703       &0.2874       &0.2875 \\
            \hline
            outdoor night1  &0 - 10     &0.3734       &0.2962       &0.2962 \\
                            &0 - 20     &0.3463       &0.3134       &0.3134 \\
                            &0 - 30     &0.2972       &0.2885       &0.2884 \\
            \hline
            indoor flying1  &0 - 10     &0.1162       &0.1133       &0.1190 \\
                            &0 - 20     &0.0691       &0.0803       &0.0595 \\
                            &0 - 30     &0.0493       &0.0646       &0.0396 \\
            \bottomrule
        \end{tabular}
        \label{tab:ablation}
        \vspace{-0.5cm}
    \end{table*}

The reviewed works employ different event-based depth estimation strategies, leveraging the high temporal resolution and sparsity of event data. We compare performance with the mapping module of Event-based Stereo Visual Odometry (ESVO) \cite{ZhouEventBasedStereo} that integrates multiple stereo-based techniques, including Generalized Time-Based Stereo (GTS) and Semi-Global Matching (SGM), with a probabilistic depth fusion framework. It models an inverse depth uncertainty and refining estimates over time to achieves robust depth estimation in high-speed and low-light conditions. Also the Event-Based Multi-View Stereo (EMVS) \cite{Rebecq18ijcv} adopts a monocular approach, reconstructing depth by back-projecting events into a DSI and using a space-sweep voting strategy. This eliminates the need for explicit feature matching, making it computationally efficient and real-time capable on standard CPUs. Lastly, we compared against a correspondence-free stereo approach that aggregates Disparity Space Images (DSIs) from multiple event cameras using fusion techniques such as the harmonic mean \cite{Ghosh_2022}. This method bypasses event matching, mitigating synchronization errors, and improving depth robustness. In contrast, our method leverages a model-based learning approach, trained on a large-scale dataset that features diverse environmental conditions and motion patterns. Unlike our reference-based models that depend on handcrafted constraints or epipolar geometry, by leveraging self-supervised learning, we remove dependencies on external annotation and preserve the asynchronous nature of event cameras.   

In table \ref{tab:performance} we report the error metrics from our model and compare them with the base methods which highlights the strengths and limitations of our approach. Our method achieves the lowest absolute mean error across all test cases, significantly outperforming the baseline methods. This demonstrates superior accuracy in estimating absolute depth values. However, while our approach excels in estimating absolute errors, it struggles with maintaining proportional depth estimation. This suggests our student-teacher training strategy is unreliable if the teacher has only been exposed to limited data variability.  


    \subsection{Ablation}
    

    This section aims to unpick the performance contributions of the SSL dataset and the training methodology we propose. To enhance the depth prediction from our self-supervised network model we perform ablation studies on two transfer learning strategies, namely fine-tuning and our proposed teacher-student training strategies.  \\
    
    \noindent
    \textit{Finetuning Training Strategy ($EvtSL \to EvtSSL$):} \\
        For this training strategy, an initial Event-Based Supervised Learning  $EvtSL$ approach is used as the backbone model. The backbone was pre-trained on the MVSEC training split. Consequently, the EvtSSL is then trained by fine-tuning the backbone model on the EvtSlowTV dataset in a self-supervised manner. This approach proves effective in predicting depth but is often constrained by the architecture of the backbone model. 

    \noindent
    \textit{Teacher-Student ( $EvtSL \to \overline{EvtSSL}$):}\\
        To allow more flexibility in designing a custom Event-Based Self-Supervised Learning EvtSSL we implemented this training strategy, that allows the main model to learn robust depth representations by distilling knowledge from a high-quality depth supervision model. Depth accuracy from the self-supervised network model progressively improves by mimicking prediction from the EvtSSL during training. In this case, $EvtSL$ is the teacher model which has been pre-trained as the backbone model. 
        
     In Table \ref{tab:ablation} we report the Root Mean Square (RMS) error with day and nighttime driving samples as well as a drone footage sequence from the MVSEC dataset. 
    The results show the relatively close values between the two self-supervised training strategies. We observe significant improvements in the self-supervised models across all sequences and metrics compared to the supervised model. The prediction error of these models is consistent despite lighting and motion challenges in the test samples. This indicates that the models can generalize well because of their exposure to a large-scale and varied dataset. 
    

    The result from the teacher-student training strategy appears to be closer to fine-tuned and more helpful with \textit{indoor\_flying} sequence because it is more varied and benefits from supervised grounding. 
     
   A reduced error within 10-meters cut-off distance with the self-supervised models reflects the models can make better predictions with reliable motion and edge. This is because closer objects appear to have more profound transformations than distant objects. This information proofs that with more constraints to accurate depth estimation. This also indicates that our depth estimation framework is effective in preserving the geometric consistency of event streams. 


\vspace{-0.2cm}
\section{Conclusion}
    \label{sec:conc}
    \vspace{-0.2cm}
    To address the challenges of limited large-scale event datasets, we introduced EvtSlowTV, a synthetic event dataset generated from real-world video sequences. This dataset provides a rich, diverse, and high-resolution event stream, enabling better generalization and robustness in depth estimation under challenging lighting and motion conditions. To assess the functionality of our dataset, we implemented a depth estimation network model that leverages a high-quality depth supervision model. Our model proves effective in preserving geometric consistency from event streams. Notably, our self-supervised training approach with the proposed contrast maximization loss function, further enhances depth accuracy, mitigating the challenges of sparse event distributions at greater distances.
    Extensive evaluations demonstrate that our method performs significantly better than existing methods, particularly in challenging environmental conditions. We hope the order of magnitude increase in training data availability helps to support future developments in the field.

{\small
\bibliographystyle{ieee_fullname}
\bibliography{egbib}
}

\end{document}